\documentclass[letterpaper]{article} % DO NOT CHANGE THIS
\usepackage{aaai25}  % DO NOT CHANGE THIS
\usepackage{times}  % DO NOT CHANGE THIS
\usepackage{helvet}  % DO NOT CHANGE THIS
\usepackage{courier}  % DO NOT CHANGE THIS
\usepackage[hyphens]{url}  % DO NOT CHANGE THIS
\usepackage{graphicx} % DO NOT CHANGE THIS
\usepackage{natbib}  % DO NOT CHANGE THIS AND DO NOT ADD ANY OPTIONS TO IT
\usepackage{caption} % DO NOT CHANGE THIS AND DO NOT ADD ANY OPTIONS TO IT
\usepackage{algorithm}
\usepackage{algorithmic}
\usepackage{amsmath,amsfonts,amssymb}
\usepackage{multirow} 
\usepackage{multicol} 
\usepackage{booktabs}
\usepackage{utfsym}
\usepackage{arydshln}
\usepackage{graphicx}
\usepackage{color} 
\usepackage{makecell}

\usepackage{newfloat}
\usepackage{listings}
\DeclareCaptionStyle{ruled}{labelfont=normalfont,labelsep=colon,strut=off} % DO NOT CHANGE THIS
\floatstyle{ruled}
\newfloat{listing}{tb}{lst}{}
\floatname{listing}{Listing}
\title{Asymmetric Reinforcing against Multi-modal Representation Bias}
\author{
    Xiyuan Gao\textsuperscript{\rm 1,2}, Bing Cao\textsuperscript{\rm 1,2}\thanks{Corresponding author}, Pengfei Zhu\textsuperscript{\rm 1}, Nannan Wang\textsuperscript{\rm 2}, Qinghua Hu\textsuperscript{\rm 1}
}
\affiliations{
    \textsuperscript{\rm 1}College of Intelligence and Computing, Tianjin University, Tianjin, 300000, China\\
    \textsuperscript{\rm 2}The State Key Laboratory of Integrated Services Networks, Xidian University, Xi'an, 710000, China
    \{gaoxiyuan, caobing, zhupengfei, huqinghua\}@tju.edu.cn, nnwang@xidian.edu.cn
}

\usepackage{bibentry}
\begin{document}

\maketitle

\begin{abstract}
The strength of multimodal learning lies in its ability to integrate information from various sources, providing rich and comprehensive insights. However, in real-world scenarios, multi-modal systems often face the challenge of dynamic modality contributions, the dominance of different modalities may change with the environments, leading to suboptimal performance in multimodal learning. Current methods mainly enhance weak modalities to balance multimodal representation bias, which inevitably optimizes from a partial-modality perspective, easily leading to performance descending for dominant modalities. To address this problem, we propose an \textbf{A}symmetric \textbf{R}einforcing method against \textbf{M}ulti-modal representation bias (\textbf{ARM}). Our ARM dynamically reinforces the weak modalities while maintaining the ability to represent dominant modalities through conditional mutual information. Moreover, we provide an in-depth analysis that optimizing certain modalities could cause information loss and prevent leveraging the full advantages of multimodal data. By exploring the dominance and narrowing the contribution gaps between modalities, we have significantly improved the performance of multimodal learning, making notable progress in mitigating imbalanced multimodal learning. Our code is available at https://github.com/Gao-xiyuan/ARM.
\end{abstract}


\section{Introduction}
Multimodal learning has emerged as a pivotal area in the field of machine learning, leveraging data from multiple sources to enhance the performance of models. This approach has been particularly transformative in applications such as image and text analysis, speech recognition, and autonomous driving, where combining visual, auditory, and textual information leads to more robust systems and makes multimodal learning an exciting frontier with significant potential \cite{whatmakes}. Despite promising yields, multimodal learning faces a critical challenge: \textit{imbalanced learning among different modalities}. In most scenarios, partial modalities, even a single modality, may dominate the learning process, leading to insufficient learning of other modalities. Some modalities may become hard to learn due to environmental interference or limited information, leading to a multimodal bias for easier-to-learn modalities \cite{greedy}, and multimodal learning may degrade to unimodal learning \cite{whatmodality}. 

\begin{figure}[t]
    \centering
    \includegraphics[width=0.9\columnwidth]{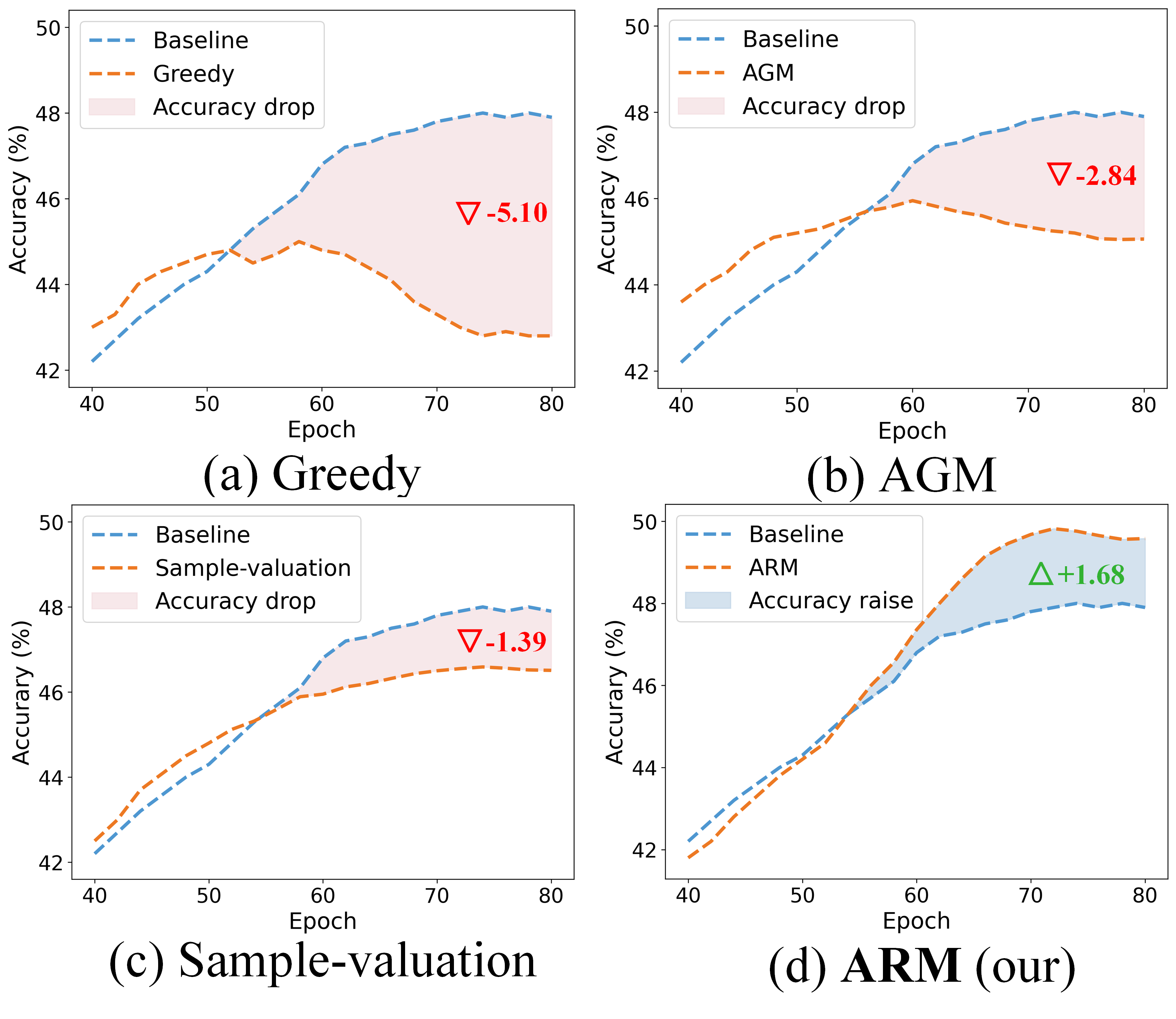}
    \caption{Accuracy curve of dominant modality compared with joint training baseline of imbalanced multimodal learning methods on Kinetics Sounds dataset. Other methods: Greedy \cite{greedy}, AGM\cite{agm}, Sample-valuation \cite{sample}.}
    
    \label{fig1}
\end{figure}

Imbalanced learning undermines the core objective of multimodal learning, which is to harness the complementary strengths of different data formats to achieve superior performance. In recent years, many extraordinary methods have been proposed to solve this problem, including canonical correlation analysis \cite{cca}, random forest \cite{rf} and ensemble learning \cite{gbt}. 
Coupled with continuously optimized large-scale datasets and algorithm innovation, deep learning methods have shown significant promise in addressing modality imbalance \cite{im-gan,im-det}. The researchers attempted to balance the multimodal learning process through methods such as gradient modulation \cite{otf,agm}, collaborative learning \cite{review}, and evaluation of modality contributions \cite{sample}. However, these methods alleviate the imbalance by improving the representation of weak modalities from the uni-modal perspective alone, ignoring the connection between modalities, and not effectively utilizing all modalities. Although some methods \cite{mla, shape} consider cross-modal learning, they approach it from late fusion or modality preservation, without fully exploring the interrelationships between modalities, which limits their potential to improve model performance. Therefore, how to balance multimodal cooperation from a multimodal perspective remains an open question. Specifically, it is still expected to be addressed to narrow the contribution gaps between modalities and enhance the joint contribution of all modalities by exploring the interaction information between them.

To this concern, we have introduced a comprehensive valuation metric to evaluate the marginal contribution of each modality and the joint contribution of all modalities during learning for each sample. Mutual information (MI) originates from information theory used to measure the correlation between two random variables \cite{mi}. It represents the amount of information one variable contains about another and copes with capturing arbitrary dependency relationships, including linear, nonlinear, and higher-order relationships. This inspires us to use MI to measure the contribution of each modality to the learning process. To fully explore interaction information between modalities, we further utilize Conditional Mutual Information (CMI) to measure the reduction in uncertainty brought by introducing additional modalities on top of a uni-modal, thereby balancing multimodal learning without modality forgetting. Based on this, we propose an asymmetric enhancement method to dynamically alleviate imbalanced multimodal learning while maintaining the performance of dominant modalities.
As shown in Fig.~\ref{fig1}, most imbalanced multimodal learning methods exhibit dominant modality-forgetting during the training process because their optimization does not pay sufficient attention to dominant modalities, failing to maintain performance on these modalities. In contrast, based on the interrelationships between modalities, our method not only reasonably reduces the contribution disparity between modalities but also enhances the performance of each modality, overcoming the modality-forgetting. The main highlights of our study are as follows:

\begin{itemize}
    \item We propose a mutual information-based valuation metric (MIV) to measure the marginal contribution of each modality and the joint contribution of all modalities in a sample with interrelation between modalities. 
    \item Based on MIV, we propose an asymmetric reinforcement framework for multimodal representation bias, which dynamically narrows the contribution gaps between modalities. By continuously focusing on the dynamically changing dominance of different modalities, we mitigate modality forgetting and enhance the overall performance.
    \item We first reveal modality contributions from a multimodal perspective, each modality makes a positive and unique contribution to the multimodal systems. Extensive experiments validated our superiority on various multimodal classification datasets against the SOTAs.
    
\end{itemize}

\section{Related Works}

\subsection{Imbalanced Multimodal Cooperation} 
Most multimodal learning often struggles with modality bias, where the dominant modality overshadows the others, leading to suboptimal performance. Recent advancements have focused on addressing this phenomenon through prototypical network \cite{pmr}, gradient modulation \cite{im-gm,otf}, and distilling knowledge \cite{disentangled,disentangled2}, dynamically weighing the importance of each modality based on task relevance or transferring knowledge from well-trained models, helping to mitigate imbalance. Evaluation methods \cite{visualwebarena,eval-mm}, especially SHAPE \cite{shape} and Sample-valuation \cite{sample} novelly encourage balanced learning by improving the optimization of worse score modalities. 
Despite these advances, challenges remain in achieving truly balanced multimodal learning, most of these methods fall short by only enhancing weaker modalities without considering the intricate relationships between them. 
In contrast, we provide an asymmetric reinforcement strategy that dynamically alleviates multimodal bias based on contribution estimation without modality forgetting.
This approach not only reduces the contribution disparity between modalities but also enhances overall multimodal cooperation, leading to improved performance across various multimodal classification datasets.

\subsection{Mutual Information in Machine Learning} 
Mutual Information (MI) has been a fundamental concept in information theory and its applications in machine learning \cite{nips-mi,icassp-mi}, which highlights the dependency between variables. In machine learning, MI is widely used for feature selection and representation learning. Early techniques \cite{fs,fs2} utilized MI to identify the most relevant features for predictive modeling, improve model performance by removing redundant or irrelevant features, and allow models to focus on the most informative features. In deep learning, MI has been instrumental in unsupervised learning and generative models \cite{nmi-seg}. Techniques like InfoGAN \cite{infogan} leverage MI to improve the quality of generated samples and the robustness of models. Some variational autoencoders mutations \cite{vae-mi} use MI to learn a latent representation that captures the underlying structure of the data while ensuring independence between latent variables. Furthermore, MI neural estimation \cite{nips-mi2} has been introduced to efficiently estimate MI between high-dimensional variables, enabling more accurate learning in complex models. Recent advancements also include using MI in knowledge distillation \cite{kd-mi} and domain adaptation \cite{da-mi}, where understanding the information flow between different domains or causal variables is crucial. Overall, mutual information continues to be a powerful tool in enhancing the capabilities of machine learning models, driving us to use mutual information to measure modal benefits. To the best of our knowledge, we for the first time utilize mutual information to handle imbalanced multimodal learning.

\begin{figure*}[htbp]
    \centering
    \includegraphics[width=0.8\linewidth]{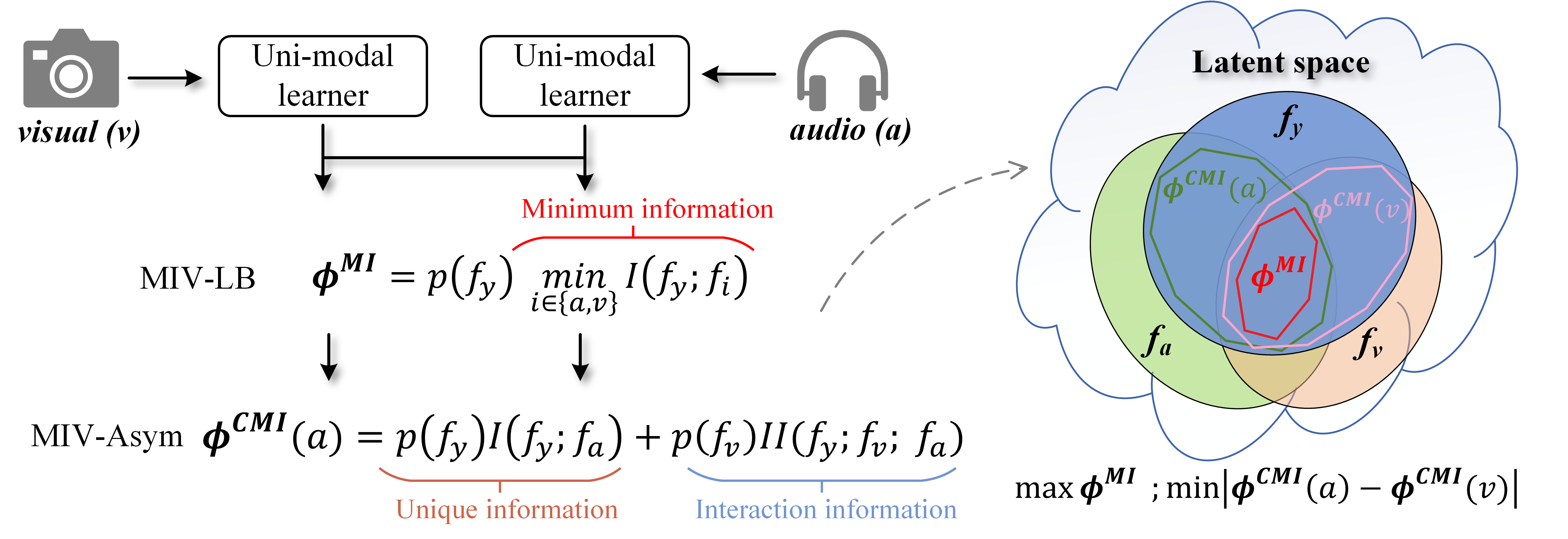}
    \caption{\textbf{Left:} The Lower Bound joint contribution (MIV-LB) of all modalities and the Asymmetric marginal contribution (MIV-Asym) of each modality are estimated by $\phi^{MI}$ and $\phi^{CMI}$, respectively, serving as the basis for asymmetric reinforcement. $f_\mathcal{Y}$ is feature-level fusion result, $p$ is the accurate production. \textbf{Right:} Representation of features in the latent space. We minimize the diversities in $\phi^{CMI}$ to balance multimodal learning while maximizing $\phi^{MI}$ to enhance multimodal performance.
    }
    \label{fig2}
\end{figure*}
\section{Methods}

\subsection{Preliminary} 
In an interactive system, we can obtain partial information about one variable \textit{X} by observing another variable \textit{Y}, thereby reducing the uncertainty of the former. The extent of this uncertainty reduction can be considered a measure of contribution and can be quantified using Mutual Information (MI). Using the basic relationship between the MI and entropy $H(\cdot)$ \cite{mi}, the algorithm for MI can be defined as the individual entropy of \textit{X}, minus the conditional entropy of \textit{X} given \textit{Y}. Following this approach, we can derive the formula for MI and Normalized MI (NMI) as follows:
\begin{align}\label{eq1}
    %\nonumber
    I(X ; Y)&= \sum_{x\in X} \sum_{y\in Y} \mathcal{P}(x, y) \log \frac{\mathcal{P}(x, y)}{\mathcal{P}(x) \mathcal{P}(y)}, 
    \\
    NMI&(X;Y)=\frac{I(X;Y)}{\sqrt{H(X)H(Y)} }. \label{eq2}
\end{align}

%where $D_{KL}$ is the Kullback-Leibler divergence. 
Considering a multimodal classification task, a sample with \textit{m} modalities is represented as $\mathcal{X} = \{ x^1, x^2, \dots, x^m \}$, which can be regarded as a multimodal pair, and $\mathit{y}$ is the ground truth label of sample $\mathcal{X}$. Denote a uni-modal encoder as $\mathcal{E}(\cdot)$, the classification head as $\mathcal{H}(\cdot)$. The feature of the \textit{i}-th modality extracted by the encoder is \textit{$f_{x^i} = \mathcal{E}(x^i)$}. When taking $\mathcal{X}$ as the input for multimodal learning, the feature-level fusion output is $f_\mathcal{Y} = \cup f_{x^i}$, $x^i \in \mathcal{X}$, the final prediction is $\mathit{\hat{y}=\mathcal{H}(f_\mathcal{Y})}$. Notably, in this multimodal classification task, the features $f_{x^i}$ of the multimodal pair $\mathcal{X}$ are fused to obtain $f_\mathcal{Y}$. Subsequently, $f_\mathcal{Y}$ is used to make the final prediction $\mathit{\hat{y}}$, and the parameters of $\mathcal{E}(\cdot)$ are optimized by backpropagation based on $\mathit{\hat{y}}$, that is, $f_\mathcal{Y}$ further applied to $f_{x^i}$. Hence, a system characterized by the mutual interaction between $f_{x^i}$ and $f_\mathcal{Y}$ is constituted.

\subsection{Valuation Metric without Modality Forget}
When the number of variables in an interactive system increases, such as in multimodal learning, where features $f_{\mathcal{X}}=\{f_{x^1}, f_{x^2}, \dots, f_{x^m}\}$ from \textit{m} modalities jointly influence the fusion result $f_\mathcal{Y}$, using mutual information can become challenging. Inspired by exhaustively decomposing in a multivariate system \cite{nonnegative}, even in cases where multiple source variables jointly influence a single variable, we can still compute the MI: $I(f_\mathcal{Y};f_{x^i})$ for each $f_{x^i} \in f_{\mathcal{X}}$ with $f_\mathcal{Y}$ separately. Notably, Eq. \eqref{eq1} is non-negative, so it has a positive contribution to learning each modality. The MI between $f_\mathcal{X}$ and $f_\mathcal{Y}$ can be expressed as:

\begin{equation}
\label{eq3}
    I(f_\mathcal{Y} ; f_{\mathcal{X}}) = \sum_{\hat{y}\in f_\mathcal{Y}} \sum_{x\in f_{\mathcal{X}}} \mathcal{P}(x, \hat{y}) \log \frac{\mathcal{P}(x, \hat{y})}{\mathcal{P}(x) \mathcal{P}(\hat{y})}, 
\end{equation}
\noindent through observing $f_\mathcal{Y}$, the distribution of $f_\mathcal{X}$ changes from 
$\mathcal{P}(x)$ to $\mathcal{P}(x|\hat{y})$, we can capture the knowledge of $f_\mathcal{X}$ after the observation, the positive contribution in Eq. \eqref{eq3} is where predicting the ground truth label $y$, that is:

\begin{equation}
\label{eq4}
    I(f_\mathcal{Y} = y ; f_{\mathcal{X}}) = \sum_{\boldsymbol{x}\in f_{\mathcal{X}}} \mathcal{P}(\boldsymbol{x}|y) \log \frac{\mathcal{P}(y \mid \boldsymbol{x})}{\mathcal{P}(y)}.
\end{equation}

\noindent \textbf{Theorem 1.} \textit{In multimodal learning with m modalities, each modality can provide a \textbf{positive} and \textbf{unique} contribution to accurate prediction. i.e., $I(f_\mathcal{Y} = y ; f_{x^i}) \neq I(f_\mathcal{Y} = y ; f_{x^j})$, for any $x^i,x^j \in \mathcal{X}, i\neq j$. Naturally, neglecting the learning of any modality will result in information loss.} (The specific theoretical proof process is provided in the Appendix.)

Based \textbf{Theorem 1}, we propose a valuation metric to measure the marginal contribution of each modality in a sample $\mathcal{X}$, i.e. $\phi(x^i)$ and further derive the joint contribution of all modalities in that sample, i.e. $\phi(\mathcal{X})$. This serves as the foundation for asymmetric enhancement.
\subsubsection{Lower bound of joint contribution $\phi(\mathcal{X})$.}
NMI between uni-modal and fused feature $NMI(f_\mathcal{Y};f_{x^i})$ can be understood as the expected contribution value of all possible predictions from $f_\mathcal{Y}$ when $f_{x^i}$ is given, and it can be expressed as Eq. \eqref{eq5}. For clarity, we use $I$ to represent $NMI$.
\begin{equation}\label{eq5}
    I\left(f_\mathcal{Y} ; f_{x^{i}}\right)=\sum_{\widehat{\mathit{y}}}^{\mathrm{N}} p\left(f_\mathcal{Y} \rightarrow \hat{y}\right) I\left(f_\mathcal{Y} ; f_{x^{i}}\right), 
\end{equation}

\noindent where $p(f_\mathcal{Y} \rightarrow \hat{y})$ represent the probability that $f_\mathcal{Y}$ makes the final prediction of class $\mathit{\hat{y}}$, \textit{N} is the number of categories. As we adopt \textit{Softmax}, $\sum_{\widehat{\mathit{y}}}^{\mathrm{N}} p\left(f_\mathcal{Y} \rightarrow \hat{y}\right) =1$. Therefore, based on the MI, the contribution of the model's accurate prediction provided by $i$-th modality can be written as:

\begin{equation}\label{eq6}
    \phi^{M I} \left(x^{i}\right)=p\left(f_\mathcal{Y} \rightarrow y\right) I\left(f_\mathcal{Y} ; f_{x^{i}}\right).
\end{equation}

Similarly, observing $j$-th modality ($j\neq i$) can also contribute to an extent that $f_\mathcal{Y}$ makes the accurate prediction $y$. Hence, the lower bound of joint contribution for all modalities in sample $\mathcal{X}$ is:

\begin{equation}\label{eq7}
    \phi^{M I}(\mathcal{X})=p\left(f_\mathcal{Y} \rightarrow y\right) \min _{i \in\{1, \ldots m\}} I\left(f_\mathcal{Y} ; f_{x^{i}}\right).
\end{equation}

It represents the minimum contribution value that each modality can provide for the model's accurate prediction. $\phi^{M I}$ has several properties: Firstly, its value range is [0,1]. Secondly, $\phi^{M I}$ is less than or equal to $I(f_\mathcal{Y};f_{x^i})$ for all $i\leq m$. Finally, in the training phase, by incorporating $\phi^{MI}$ into the loss function and using gradient descent to maximize $\phi^{MI}$, thus each iteration moves towards increasing mutual information, ensuring the convergence of the lower bound.

\subsubsection{Estimating marginal contribution $\phi(x^i)$.} Although we defined the lower bound joint contribution of sample $\mathcal{X}$, the interrelationships between modalities are ignored, which prevents us from fully leveraging the advantages of multimodal learning. 
As one would hope, given the presence of variable $Z$, the impact of introducing an additional variable $Y$ on $X$ can be measured using Conditional Mutual Information (CMI). The formulas for CMI and Normalized CMI (NCMI) are as follows:
\begin{align}\label{eq8}
    \nonumber
    CMI {\scriptstyle (X;Y|Z)}&=  \sum_{x\in X}^{} \sum_{y\in Y}^{} \sum_{z\in Z}^{}  \mathcal{P}(x,y,z)\log_{}{} \frac{\mathcal{P}(x,y|z)}{\mathcal{P}(x|z)\mathcal{P}(y|z)}, 
    \\
    &=\mathbb{E}_Z D_{KL} \left[\mathcal{P}(x, y|z) \| \mathcal{P}(x|z) \mathcal{P}(y|z)\right] \\
    NCMI {\scriptstyle (X;Y}&{\scriptstyle|Z)}=\frac{CMI(X;Y|Z)}{\sqrt{H(X|Z)H(Y|Z)}}. \label{eq9}
\end{align}

In a complete modality set $\mathcal{X}$, when we choose the $x^i$ and $x^j$ to calculate MI with the fusion result separately, the mutual information of $f_{x^i}$ is $I(f_\mathcal{Y};f_{x^i})$, and the conditional mutual information of $f_{x^j}$ given $f_{x^i}$ is $I(f_\mathcal{Y};f_{x^j}|f_{x^i})$, vice versa. Consequently, the change in contribution value that modality $x^j$ causes to modality $x^i$ is the Interaction Information (II):
\begin{equation}\label{eq10}
    II(f_{\mathcal{Y}};f_{x^j};f_{x^i})=I(f_{\mathcal{Y}};f_{x^j})-{\scriptstyle NCMI}(f_{\mathcal{Y}};f_{x^j}|f_{x^i}).
\end{equation}

With Eq.~\eqref{eq10}, we can estimate the marginal contribution of $x^i$ based on considering all modalities as follows:
\begin{align} \label{eq11}
    \nonumber \phi^{CMI}(x^{i}) &=  p\left(f_\mathcal{Y} \rightarrow y\right) I\left(f_\mathcal{Y} ; f_{x^{i}}\right) \\
    &+ \sum_{j\neq i}^{m} p(f_{x^j}\rightarrow y)II(f_{\mathcal{Y}};f_{x^j};f_{x^i}),
\end{align}

\noindent where $p(f_{x^j}\rightarrow y)$ can be regarded as a dynamic modality-specific weight of $j$-th modality, which can heighten the model's robustness \cite{quantifying}. Furthermore, the joint contribution of the complete modality set from sample $\mathcal{X}$ can be expressed as:

\begin{equation}\label{eq12}
    \phi^{CMI}(\mathcal{X}) = \frac{1}{m} \sum_{i=1}^{m}\phi^{CMI}(x^{i}).
\end{equation}

$\phi^{CMI}$ has several advantages: Firstly, it considers the impact of each modality from sample $\mathcal{X}$, ensuring that there is no modality omission during learning. Secondly, its value range is [0, m], allowing it to be flexibly incorporated into loss functions or regularization as an optimization technique. Finally, averaging reasonably reflects the salient characteristics of the overall modalities, preventing the landslide victory of certain modalities while suppressing the occurrence of outliers.

\subsection{Asymmetric Reinforcement Strategies}
\subsubsection{Dynamic Feature-level Fusion.} Considering real-world factors, due to the primacy effect, the effect of the first term in Eq. \eqref{eq11} will be amplified. In other words, $\phi^{CMI}(x^{i})$ reflects the importance to accurate prediction of $i$-th modality. We can use this as the specific-modal fusion weight during the fusion phase. Generally, higher $\phi^{CMI}(x^{i})$ values represent more positive impacts on the model, thus the Fusion Weight (FW) of $i$-th modality can be denoted as:
\begin{equation}\label{eq13}
    FW^{i}=\frac {\phi^{CMI}(x^i) }{\phi^{CMI}(\mathcal{X})},
\end{equation}

\noindent where $FW^i$ works during the training phase and will take effect in the next epoch.

\subsubsection{Balanced Min-Max Loss.} Examining the expression of Eq. \eqref{eq7}, \eqref{eq12}, it is evident that $\phi^{MI}$ and $\phi^{CMI}$ represent the minimum contribution and comprehensive contribution that complete modalities for the model's accurate prediction, respectively. For the former, maximizing $\phi^{MI}$ enables the model to learn the most beneficial aspect of each modality for accurate prediction, and for the latter, we can use the Mean Absolute Error (MAE) to minimize $MAE(\phi^{CMI})$, thereby narrowing the marginal contribution gap between modalities.
\begin{align}
    \label{eq14}\mathcal{L}_{\phi^{MI}} &=1-\phi^{MI}(\mathcal{X} ), \\
    \label{eq15}\mathcal{L}_{\phi^{CMI}} &=\frac{\sum_{i=1}^{m}\left |\phi^{CMI}(x^i)-  \phi^{CMI}(\mathcal{X})\right |  }{\phi^{CMI}(\mathcal{X})}.
\end{align}

It should be noted that Eq. \eqref{eq14} cannot directly participate in the gradient backward process of gradient descent optimization since the min function is not globally differentiable. To this end, we use smooth approximation \cite{min} to make it differentiable:
\begin{equation} \label{eq16}
    \min _{i \in\{1, \ldots m\}} I^i=\max _{i \in\{1, \ldots m\}}(-I^i)\approx \log (\sum_{i=1}^{m} e^{-I^i} ).
\end{equation}

The overall loss function of ARM is formulated as Eq. \eqref{eq17}, where $\mathcal{L}_{CE}$ denotes the cross-entropy loss, $\lambda_1$ and $\lambda_2$ are trade-off parameters.
\begin{equation}\label{eq17}
    \mathcal{L}=\mathcal{L}_{CE}+\lambda _{1}\mathcal{L}_{\phi^{MI}}  + \lambda _{2}\mathcal{L}_{\phi^{CMI}}. 
\end{equation}

\subsubsection{Dynamic Sample-level Re-sample.} Following the analysis in \textbf{Theorem 1} and specific theoretical in \cite{sample}, enhancing the discriminative ability of lower-contribution modality can expand its contribution. We propose to resample all modalities of lower joint contribution sample $\mathcal{X}$ more frequently during training. After each modality valuation by MIV, we can dynamically determine the re-sampling frequency with $\phi^{CMI}$ to enhance contribution, where re-sample frequency of sample $\mathcal{X}$ is:
\begin{equation}\label{eq18}
    s(\mathcal{X})= \mathcal{F}_s(\phi^{CMI}(\mathcal{X})),
\end{equation}

\noindent where $\mathcal{F}_s$ is a monotonically decreasing function, the lower-contribution sample $\mathcal{X}$ is re-trained with a resample frequency inversely proportional to its joint contribution. It is worth noting that different from \cite{sample}, our resampling strategy is from a multimodal perspective, which dynamically adjusts the sampling frequency of all modalities in $\mathcal{X}$. This ensures that no information is lost during training, while the loss function $\mathcal{L}_{\phi^{MI}}$ guarantees targeted learning for lower-contribution modalities.

\section{Experiments}

\subsection{Datasets and Implementation Details}
\noindent\textbf{Kinetic Sounds (KS)} \cite{ks} is a specifically designed action recognition dataset for research in audio-visual learning, particularly focusing on the relationship between actions and corresponding sounds. KS is composed of YouTube videos; all videos are cropped to within 10 seconds around the action. KS includes approximately 23k video clips with 31 categories.

\noindent\textbf{UCF-51} is a subset of UCF-101 \cite{ucf101} with two modalities, RGB and optical flow, containing 6,845 video clips across 51 diverse action categories. Mostly sourced from YouTube, it features varying conditions such as different camera angles and lighting, making it challenging and realistic for real-world applications.

\noindent\textbf{UPMC Food-101} \cite{food} is a comprehensive dataset for food recognition, consisting of 101,000 images accompanied by corresponding texts across 101 food categories. Each category includes 750 images for training and 250 images for testing.

\subsubsection{Implementation Details.} Unless otherwise specified, ResNet-18 is used as the backbone in the experiments and trained from scratch. Encoders used for UCF-51 are ImageNet pre-trained. 
For Food-101, a ViT-based model is used as the vision encoder, and a BERT-based model is used as the text encoder by the pre-trained. Before modality valuation, a warm-up stage is employed for all experiments. During training, we use Stochastic Gradient Descent (SGD) with a batch size of 64. We set the initial learning rate, weight decay, and momentum parameters to $10^{-3}$, $5\times10^{-4}$, and $0.9$, respectively. The experiments are conducted on Huawei Atlas 800 Training Server with CANN and NVIDIA 4090 GPU. More details of implementation and experiment analysis are provided in the Appendix.

\subsection{Comparison with Imbalanced Multimodal Learning Methods}

\begin{table}[t]
    \centering
    \begingroup
    \setlength{\tabcolsep}{4pt} % Default value: 6pt
    \renewcommand{\arraystretch}{0.7} % Default value: 1

    \begin{tabular}{ l  c  c }
    \toprule
    \textbf{Model} & KS (Acc.) & UCF-51 (Acc.)\\ \midrule \midrule
    Concatenation	&$59.61$	&$68.23$\\
    Summation &$59.53$ &$67.62$ \\ \midrule
    OGM-GE \scriptsize{(CVPR 2022)}	&$60.70$	&$71.66$\\
    Greedy \scriptsize{(ICML 2022)}	&$59.86$	&$71.53$\\
    QMF \scriptsize{(ICML 2023)}	&$63.78$	&$73.48$\\
    PMR \scriptsize{(CVPR 2023)}	&$63.86$	&$74.80$\\
    Sample-val.  \scriptsize{(CVPR 2024)}	&\textcolor{blue}{$65.33$}	&$75.12$\\
    Modality-val.  \scriptsize{(CVPR 2024)}	&$65.10$	&$74.39$\\
    MLA \scriptsize{(CVPR 2024)}	&$65.21$	&\textcolor{red}{$76.01$}\\ \midrule
    \textbf{ARM}	&\textcolor{red}{$66.52$}	&\textcolor{blue}{$75.60$}\\ \bottomrule
 
    \end{tabular}
    \endgroup
    
    \caption{Accuracy of imbalanced multimodal learning methods, where red and blue indicate the \textcolor{red}{best}/\textcolor{blue}{runner-up} performance. Results are reported in percentage ($\%$).}
    \label{tab_imb}
\end{table}

\begin{figure}[t]
    \centering
    \includegraphics[width=0.9\linewidth]{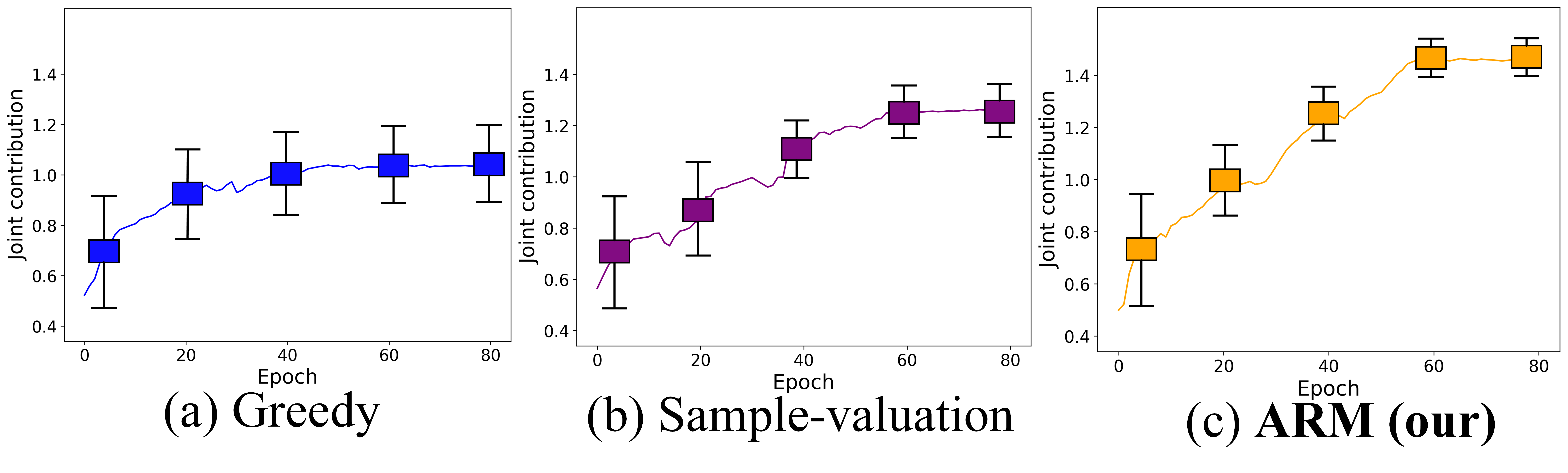}
    \caption{
    Comparison of the narrowing trend of uni-modality contribution gaps on the UCF-51 dataset.}
    \label{fig_gap}
\end{figure}

In this section, we compared ARM with advanced imbalanced multimodal learning  methods to answer  
\textbf{Q1:} \textit{How does ARM narrow the modality contribution gap?}

Fig.~\ref{fig_gap} illustrates the trend of narrowing contribution gaps across different methods. Traditional methods, like Greedy, have shown limited improvement in closing the contribution disparity between modalities, with only a slight narrowing in the contribution gap as training progresses. Sample-valuation demonstrates more consistent shrink, yet the gap remains noticeable across epochs. In contrast, ARM achieves a marked and consistent reduction in modality contribution gaps, indicating a more balanced learning process. This consistent improvement shows ARM’s ability to maintain equitable contribution from all modalities, which is crucial for robust multimodal learning.

Table \ref{tab_imb} further reinforces this conclusion. ARM consistently outperforms other state-of-the-art methods, i.e., Greedy \cite{greedy}, OGM-GE \cite{otf}, QMF~\cite{qmf}, PMR \cite{pmr}, Sample-valuation, Modality-valuation \cite{sample}, and MLA \cite{mla}, achieving the competitive accuracy scores of 66.52$\%$ and 75.60$\%$, respectively. 
Other approaches, like QMF and PMR, show decent performance but still fall short in balancing modality contributions, leading to suboptimal accuracy. 
Due to the different design focus, MLA performs better in handling temporal optical flow data in the UCF-51 dataset.
Sample-valuation achieves competitive results but cannot match the balance achieved by ARM, which is evident from the joint contribution trends shown in Fig.~\ref{fig_gap}.
The advantage of ARM lies in its dual focus: minimizing modality imbalances while maximizing overall performance. By effectively narrowing the contribution gaps between modalities, ARM prevents any single modality from dominating or being neglected, leading to a more cohesive and effective multimodal representation.

\subsection{Comparison with Multimodal Fusion Methods}
\begin{table}[t]
    \centering
    \begingroup
    \setlength{\tabcolsep}{1pt} % Default value: 6pt
    \renewcommand{\arraystretch}{0.7} % Default value: 1

    \begin{tabular}{ l  c  c }
    \toprule
    \textbf{Model} & KS (Acc.)& Food-101 (Acc.)\\ \midrule \midrule

    Concatenation	&$59.61$	&$82.38$\\
    Summation	&$59.53$	&$82.63$\\ \midrule
    Decision fusion	&$60.12$	&$83.71$\\
    FiLM \scriptsize{(AAAI 2018)}	&$59.33$	&$82.34$\\
    BiGated \scriptsize{(AAAI 2018)}	&$60.79$	&$86.71$\\
    Dynamic Fusion \scriptsize{(CVPR 2023)} 	&$63.21$	&$90.83$\\
    PMF \scriptsize{(ICCV 2023)}	&$64.33$	&\textcolor{blue}{$91.56$}\\
    TransFusion \scriptsize{(ICLR 2024)}	&\textcolor{blue}{$65.40$}	&$91.22$\\ \midrule
    \textbf{ARM}	&\textcolor{red}{$66.52$}	&\textcolor{red}{$93.36$}\\
    \bottomrule
    \end{tabular}
    \endgroup
    %\vspace{-3pt}
    \caption{Comparison with multimodal fusion methods.}

    \label{tab_fusion}

\end{table}

Table \ref{tab_fusion} compares the performance of various multimodal fusion methods on two datasets to answer \textbf{Q2:} \textit{Can the proposed modules (e.g., dynamic feature-level fusion) effectively improve performance?} 

Concatenation and Summation are baseline methods that simply merge the feature vectors, yielding moderate performance. 
More advanced techniques such as FiLM~\cite{film} and BiGated \cite{bigate} introduce interaction between modalities through modulation or gating mechanisms, resulting in eligible accuracy compared with the baseline. Dynamic Fusion \cite{dynamic} incorporates adaptive fusion strategies, which inspire our dynamic feature-level fusion, adjusting how the modalities are combined during inference, which leads to substantial improvements, especially on the Food-101 dataset.

Among the recently proposed methods, PMF \cite{pmf} and TransFusion \cite{transfusion} demonstrate the power of more sophisticated fusion techniques. PMF achieves strong performance by effectively managing modality-specific features. TransFusion, a transformer-based model, further refines this by better capturing the complex interactions between modalities, achieving runner-up results. Our ARM outperforms all other models on both datasets, achieving 66.52$\%$ accuracy on KS and 93.36$\%$ on Food-101, which is a significant improvement, particularly evident on the KS dataset, where it exceeds the runner-up by over 1 percentage point. ARM's success is attributed to its advanced asymmetric reinforcement strategy, which dynamically balances the learning from each modality, preventing the model from being biased toward the dominant modality. This ensures that both audio and visual information are utilized effectively, leading to superior performance in challenging multimodal tasks. Compared with the competing methods, ARM's ability to maintain high accuracy across different datasets demonstrates its robustness and adaptability, making it a standout choice for multimodal fusion tasks.

\subsection{Analysis of Modality Forget \& Multimodal Cooperation}

\begin{table*}[ht]
    \centering
    \begingroup
    \setlength{\tabcolsep}{9pt} % Default value: 6pt
    \renewcommand{\arraystretch}{0.7} % Default value: 1

    \begin{tabular}{c c | c  c | c  c |c c  c | c }
    \toprule
    \multicolumn{2}{c}{Dataset} & Conact. &Sum &BiGated &PMF &QMF &Sample &MLA &\textbf{ARM} \\ \midrule \midrule
    \multirow{1}[6]{*}{KS} & ($\star$) Audio &$47.35$ &$46.21$ &\color{gray}{$44.11$ ($\downarrow$)} &\color{gray}{$45.82$ ($\downarrow$) }&$47.56$ &\textcolor{gray}{$46.02$ ($\downarrow$)} &\textcolor{blue}{$49.20$} &\textcolor{red}{$\textbf{$49.95$}$} \\
    &Video &$23.65$ &$22.78$ &\color{gray}{$22.08$ ($\downarrow$)} &$25.65$ &$36.82$ &\textcolor{blue}{$42.67$} &$41.30$ &\textcolor{red}{\textbf{$44.86$}} \\
    &Mutli &$59.61$ &$59.53$ &$60.79$ &$64.33$ &$63.78$ &\textcolor{blue}{$65.33$} &$65.21$ & \textcolor{red}{\textbf{$66.52$}}\\
    
    \midrule
    \multirow{1}[6]{*}{UCF-51} & ($\star$) RGB &$60.13$ &$59.80$ &\color{gray}{$57.39$ ($\downarrow$)} &\color{gray}{$58.13$ ($\downarrow$)} &\color{gray}{$56.20$ ($\downarrow$)} &\color{gray}{$57.01$ ($\downarrow$)}&\textcolor{red}{\textbf{$64.81$}} &\textcolor{blue}{$63.29$} \\
    & OF &$29.62$ &$28.81$ &\color{gray}{$25.67$ ($\downarrow$)} &$36.21$ &$40.51$ &\textcolor{blue}{$42.33$} &$41.26$ &\textcolor{red}{\textbf{$43.19$}} \\
    & Mutli &$68.23$ &$67.62$ &$70.87$ &$72.09$ &$73.48$ &$75.12$ &\textcolor{red}{\textbf{$76.01$}} &\textcolor{blue}{$75.60$} 		 \\
    \midrule
    
    \multirow{1}[6]{*}{Food-101} &Image &$30.85$ &$31.66$ &$48.87$ &$59.21$ &$66.39$ &\textcolor{red}{\textbf{$73.49$}} &$71.58$ &\textcolor{blue}{$72.36$} \\
    &($\star$) Text 
    &$81.68$ &$80.84$ &\color{gray}{$78.51$ ($\downarrow$)} &\color{gray}{$79.66$ ($\downarrow$)} &$82.10$ &$84.43$&\textcolor{blue}{$86.42$} &\textcolor{red}{\textbf{$86.86$}} \\
    & Mutli &$82.38$ &$82.63$ &$86.71$ &$91.56$ &$91.67$ &$90.85$ &\textcolor{blue}{$93.31$} &\textcolor{red}{\textbf{$93.36$}}  \\
    \bottomrule
    \end{tabular}
    \endgroup
    
    \caption{Comparison results on audio-video, RGB-optical flow, and image-text datasets. The performance of a single modality and the results of combining all modalities ("multiple") are listed. $\star$ denotes the dominant modality and $\downarrow$ indicates a performance drop compared with Concatenation or Sum baseline.}
    \label{tab_forget}
\end{table*}

\begin{figure}[t]
    \centering
    \includegraphics[width=0.9\linewidth]{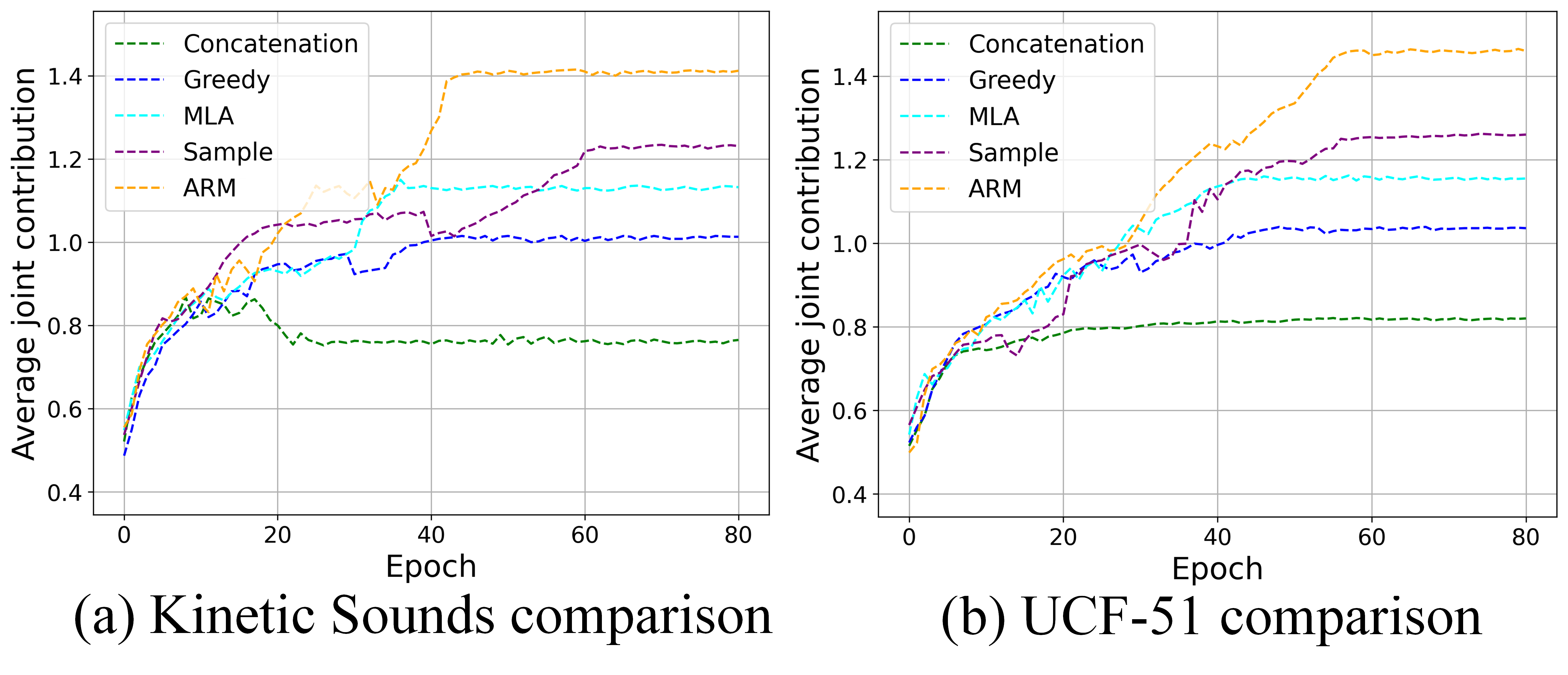}
    \caption{Average joint contribution of all modalities overall training samples during training for Greedy, MLA, Sample-valuation, and our ARM on the KS and UCF-51 datasets.}
    \label{fig_line}
\end{figure}

We report the results of a single modality and a combination of all modalities and further display the improvement of multimodal cooperation to answer 
\textbf{Q3:} \textit{Compared to prior multimodal learning approaches, can ARM overcome modality forget and optimize multimodal cooperation?} 
\subsubsection{Modality Forget.} Table \ref{tab_forget} compares the performance of various models across multiple datasets, highlighting results for different modalities and their multimodal cooperation. Several models in the comparison exhibit a notable modality forget phenomenon, where optimizing for one weaker modality leads to a decrease in the performance of the dominant modality and achieves suboptimal results in the overall multimodal performance. For instance, on the KS dataset, models like BiGated and PMF show significant drops in performance for the visual modality compared to the audio, which in turn negatively impacts their multimodal accuracy. This trend is also observed on UCF-51, where models fail to balance the learning of RGB and optical flow modalities, leading to lower overall performance. The Sample-valuation model also shows a considerable drop across both visual and textual modalities on Food-101, which further highlights the issue of modality forgetting.
Our proposed ARM consistently outperforms other models across all datasets, achieving the highest accuracy in both single and multimodal scenarios. Notably, ARM excels in preventing modality forgetting, as demonstrated by its superior performance across different modalities and their combinations.

\subsubsection{Multimodal
Cooperation.} Fig.~\ref{fig_line} illustrates the progression of multimodal joint contributions over epochs compared to Concatenation baseline for different models. The performance of the other methods indicates a relatively slower and less stable increase in the multimodal joint contribution over time. Greedy demonstrates some improvement but plateaus early, indicating that it struggles to maintain steady enhancement of multimodal cooperation. Sample and MLA show better performance than Concatenation and Greedy but still fall short compared to ARM, as they are unable to fully exploit the joint potential of multimodal learning. ARM exhibits a consistent and substantial increase in the multimodal average contribution, the chart shows that ARM not only achieves a higher overall contribution but also demonstrates a stable and continuous growth trend, indicating its robustness in learning and integrating information from various modalities.

ARM's success can be attributed to its dynamic asymmetric reinforcement strategy, which effectively balances the learning contributions from each modality based on their importance. 
By dynamically adjusting the contribution of each modality based on their importance and interaction with others, ARM ensures that no single modality dominates at the expense of others and allows ARM to maximize the joint contribution of all modalities, leading to superior performance in multimodal learning.

\subsection{The Effectiveness of Loss Function}
\begin{table}[t]
    \centering
    \begingroup
    \setlength{\tabcolsep}{5pt} % Default value: 6pt
    \renewcommand{\arraystretch}{0.7} % Default value: 1

    \begin{tabular}{ c  c  c c c c}
    \toprule
    $\mathcal{L}_{CE}$ & $\mathcal{L}_{\phi^{MI}}$ & $\mathcal{L}_{\phi^{CMI}}$ &KS &UCF-51 &Food-101 \\ \midrule \midrule
    \checkmark & & & $63.88$ &$70.03$ &$88.52$\\
    \checkmark &\checkmark & & $64.20$ &\textcolor{blue}{$73.56$} &\textcolor{blue}{$91.25$}\\
    \checkmark & &\checkmark & \textcolor{blue}{$65.19$} &$72.10$ &$89.78$\\
    \checkmark &\checkmark &\checkmark & \textcolor{red}{\textbf{$66.52$}} &\textcolor{red}{\textbf{$75.60$}} &\textcolor{red}{\textbf{$93.36$}} \\
    \bottomrule
    \end{tabular}
    \endgroup
    \caption{Ablation study of loss function.}
    \label{tab_loss}
\end{table}

This section answers the question: \textbf{Q4:} \textit{Does our proposed Balanced Min-Max loss progress as expected?}

\begin{figure}[t]
    \centering
    \includegraphics[width=0.9\linewidth]{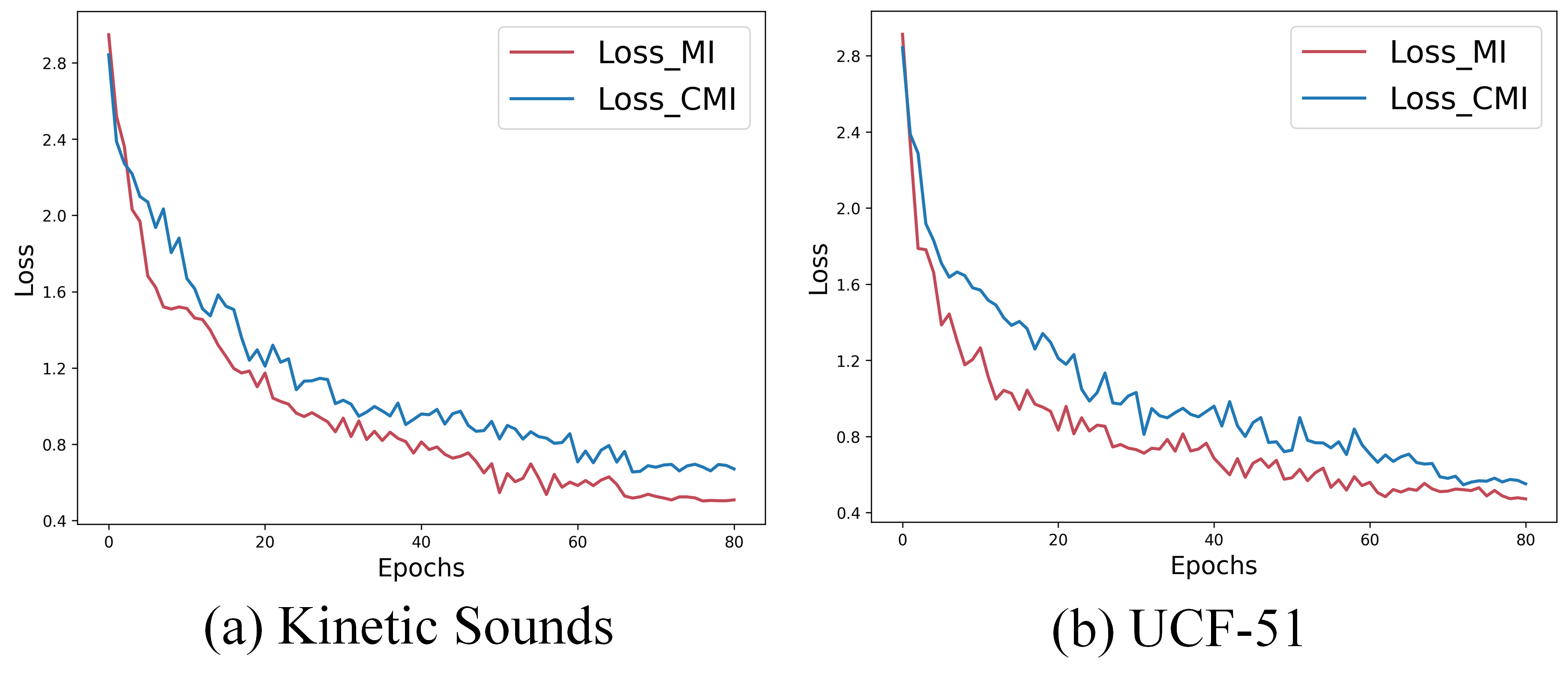}
    \caption{Curve of Balanced Min-Max Loss: the values are obtained from 5 training processes with the same initiations.}
    \label{fig_loss}
\end{figure}

Fig.~\ref{fig_loss} demonstrates the effectiveness of the proposed $\mathcal{L}_{\phi^{MI}}$ and $\mathcal{L}_{\phi^{CMI}}$ in improving overall multimodal performance and alleviating imbalanced learning between modalities, respectively. The loss curves for both the KS and UCF-51 datasets show that incorporating the Balanced Min-Max loss consistently improves the overall model performance by ensuring better multimodal cooperation, leading to faster convergence and lower loss values.
Table \ref{tab_loss} further validates these observations with an ablation study. When only the $\mathcal{L}_{\phi^{MI}}$ is added, there is a noticeable increase in accuracy compared to the baseline (row 1). Additionally, the inclusion of the $\mathcal{L}_{\phi^{CMI}}$ specifically addresses modality imbalance by reducing the learning disparity between modalities. This is particularly important in scenarios where dominant modalities may overshadow weaker ones. The combined use of both $\mathcal{L}_{\phi^{MI}}$ and $\mathcal{L}_{\phi^{CMI}}$ achieves the best performance, demonstrating that our approach not only enhances overall accuracy but also maintains balanced contributions across all modalities. This synergy between the two losses highlights the strength of our method in multimodal learning.

\section{Conclusion}
In this paper, we introduce a valuation metric to evaluate the marginal contributions of different modalities and the joint contributions of all modalities in a sample with a theoretical analysis of mutual information. Based on this, an asymmetric enhancement method named ARM is proposed to improve imbalanced multimodal learning while preventing modality forgetting. 
This provides a potential approach for balancing multimodal learning in real-world applications. 
Besides, there are some further discussions.

\noindent \textbf{Universality of Mutual Information.} Our method calculates mutual information after feature extraction, and the data dimension is lower, so we can directly calculate the marginal distribution and joint distribution. But when processing continuous data, discretization or kernel density estimation methods are required. These methods are relatively complex to implement and may lead to different results.

\noindent \textbf{Natural Conflict in Multimodal.} Multimodal data may contain some inherent conflicts. 
For example, for an RGB-Infrared sample \textit{person} in foggy environments, two modalities may make vastly different predictions. Although ARM copes with mitigating modality conflicts by reducing the impact of modalities with incorrect predictions, it does not fundamentally resolve such conflicts. Therefore, it is expected to consider this natural conflict in the future work.

\section{Acknowledgments}
This work was sponsored by National Science and Technology Major Project (No. 2022ZD0116500), National Natural Science Foundation of China (No.s 62476198, 62436002, U23B2049, 62222608, 62106171, and 61925602), Tianjin Natural Science Funds for Distinguished Young Scholar (No. 23JCJQJC00270), the Zhejiang Provincial Natural Science Foundation of China (No. LD24F020004), and CCF-Baidu  Open Fund. This work was also sponsored by CAAI-CANN Open Fund, developed on OpenI Community.
%\bigskip

\bibliography{reference}

\newpage
\section{Appendix}
In Appendix, we provide more details, proofs and experiments, encompassing the following:
\begin{itemize}
\item A theoretical proof of Theorem 1, with the help of mutual information, we have proved the indispensability of each modality. The proof is elaborated in \textit{Appx. A}.

\item More details of experiment implementation, including data processing, experiment setting and algorithm flow, as detailed in \textit{Appx. B}.

\item Additional experimental comparisons, including more ablated experiment on ARM, more discussion on dynamic sample-level re-sample strategy, more comparisons with other MI-based methods, and the improvement of ARM on other fusion methods, etc, as presented in \textit{Appx. C}.
\end{itemize}

\section{A. Proof}

\noindent \textbf{Theorem 1.} \textit{In multimodal learning with m modalities, each modality can provide a \textbf{positive} and \textbf{unique} contribution to accurate prediction. i.e., $I(f_\mathcal{Y} = y ; f_{x^i}) \neq I(f_\mathcal{Y} = y ; f_{x^j})$, for any $x^i,x^j \in \mathcal{X}, i\neq j$. Naturally, neglecting the learning of any modality will result in information loss.}
\vspace{3mm}

\noindent \textit{Proof.} 
Let us consider 3 finite non-empty feature sets of a sample $\mathcal{X}$ with $m$ modalities: \textbf{\textit{A}}=$\{f_{x^1}, f_{x^2}, \dots, f_{x^m}\}$, \textbf{\textit{B}}=$\{f_{x^1}, f_{x^2}, \dots, f_{x^n}\}$, and \textbf{\textit{C}}=$\{f_{x^{n+1}}, f_{x^{n+2}}, \dots, f_{x^m}\}$, $n$\textless$m$, $A = B \cup C$, jointly affecting \textbf{\textit{Y}=$f_{\mathcal{Y}}$}. Under the condition of ensuring accurate prediction, that is, \textbf{\textit{Y}} = $y$, we have Eq. \eqref{eq}.

Consequently, the positive contribution provided by set \textbf{\textit{A}} must be greater than that of set \textbf{\textit{B}}. Neglecting the learning of any modality will result in the loss of positive information. In other words, each modality in a sample contains a unique positive impact. This prompts us to optimize imbalanced multimodal cooperation without abandoning any modality.

\section{B. More Details}

\subsection{Implementation Details.} 
For the KS dataset, the network was trained for 80 epochs using the SGD optimizer with a momentum of 0.9, a learning rate of 0.001, and a weight decay of 0.0005. The batch size was set to 64. All videos in the KS dataset were resized to have a short edge of 256 pixels, and the sampling frequency was set to one frame per second. For the UCF-51 dataset, the encoders were initialized with ImageNet pre-trained weights, and the network was trained with an initial learning rate of 0.0005 and a batch size of 16. For the Food-101 dataset, the AdamW optimizer was used with a learning rate of 0.0001, and the backbone was also pre-trained on ImageNet. A warm-up phase was employed for all experiments, with the warm-up duration set to 10 epochs. All other hyperparameters were set to their default values as defined in PyTorch.
\begin{figure}[t]
    \begin{align*} \label{eq}
    \nonumber I(Y&=y ; \boldsymbol{A})-I(Y=y ; \boldsymbol{B}) \\
    \nonumber =&\sum_{\boldsymbol{a}} \mathcal{P}(\boldsymbol{a} \mid y) \log \frac{\mathcal{P}(y, \boldsymbol{a})}{\mathcal{P}(y) \mathcal{P}(\boldsymbol{a})} \\
    &-\sum_{\boldsymbol{b}} \mathcal{P}(\boldsymbol{b} \mid y) \log \frac{\mathcal{P}(y, \boldsymbol{b})}{\mathcal{P}(y) \mathcal{P}(\boldsymbol{b})} \\
    \nonumber =&\sum_{\boldsymbol{b}} \sum_{\boldsymbol{c}} \mathcal{P}(\boldsymbol{b}, \boldsymbol{c} \mid y) \log \frac{\mathcal{P}(y, \boldsymbol{b}, \boldsymbol{c})}{\mathcal{P}(y) \mathcal{P}(\boldsymbol{b}, \boldsymbol{c})} \\ \tag{19}
    &-\sum_{\boldsymbol{b}} \sum_{\boldsymbol{c}} \mathcal{P}(\boldsymbol{b}, \boldsymbol{c} \mid y) \log \frac{\mathcal{P}(y, \boldsymbol{b})}{\mathcal{P}(y) \mathcal{P}(\boldsymbol{b})} \\ 
=&\sum_{\boldsymbol{b}} \sum_{\boldsymbol{c}} \mathcal{P}(\boldsymbol{b}, \boldsymbol{c} \mid y) \log \frac{\mathcal{P}(\boldsymbol{b})\mathcal{P}(\boldsymbol{b})\mathcal{P}(y \mid \boldsymbol{b})\mathcal{P}(\boldsymbol{c} \mid \boldsymbol{b},y)}{\mathcal{P}(\boldsymbol{c}, \boldsymbol{b}) \mathcal{P}(y , \boldsymbol{b})} \\
\nonumber =&\sum_{\boldsymbol{b}} \sum_{\boldsymbol{c}} \mathcal{P}(\boldsymbol{b}, \boldsymbol{c} \mid y) \log \frac{\mathcal{P}(\boldsymbol{b}) \mathcal{P}(\boldsymbol{c} \mid \boldsymbol{b}, y)}{\mathcal{P}(\boldsymbol{b}, \boldsymbol{c})} \\
=&\sum_{\boldsymbol{b}} \sum_{\boldsymbol{c}} \mathcal{P}(\boldsymbol{b}) \mathcal{P}(\boldsymbol{c} \mid \boldsymbol{b}, y) \log \frac{\mathcal{P}(\boldsymbol{c} \mid \boldsymbol{b}, y)}{\mathcal{P}(\boldsymbol{c} \mid \boldsymbol{b})} \\
\nonumber =&\sum_{\boldsymbol{b}} \mathcal{P}(\boldsymbol{b}) \sum_{\boldsymbol{c}} \mathcal{P}(\boldsymbol{c} \mid \boldsymbol{b}, y) \log \frac{\mathcal{P}(\boldsymbol{c} \mid \boldsymbol{b}, y)}{\mathcal{P}(\boldsymbol{c} \mid \boldsymbol{b})} \\
\nonumber =&\medspace \mathbb{E}_B D_{KL}\left[\mathcal{P}(\boldsymbol{c} \mid \boldsymbol{b}, y) \| \mathcal{P}(\boldsymbol{c} \mid \boldsymbol{b})\right] \geq 0 
\end{align*}

\end{figure}

\subsection{Algorithm Details.}

\begin{algorithm}[t]
\setlength {\belowcaptionskip} {-5mm}
\caption{Asymmetric reinforcement strategies}
\label{alg:algorithm}

\begin{algorithmic}[1] %[1] enables line numbers
\REQUIRE  Original training dataset $\mathcal{D}$, training dataset with re-sample $\mathcal{D}^{rs}$, number of modalities $m$, loss fuction $\mathcal{L}_{total}$, model parameters
$\theta$, training epoch $T$, warm-up epoch $F$. \\
\FOR{$t=0, \dots, T-1$}
\STATE Initialize $\mathcal{D}^{rs}=\mathcal{D}$;
\FOR{each sample $\mathcal{X}= \{ x^1, x^2, \dots, x^m \}$ in $\mathcal{D}^{rs}$}
\STATE Valuate multimodal joint contribution $\phi^{MI}(\mathcal{X}), \phi^{CMI}(\mathcal{X})$ with Eq. (7), (12);
\IF{$t<F$}
\STATE Compute the loss $\mathcal{L}_{total}$ following Eq. (17);
\STATE Update parameters $\theta$ with dataset $\mathcal{D}^{rs}$;
\ELSE
\STATE Dynamic feature fusion with Eq. (13);
\STATE Get re-sample frequency $s(\mathcal{X})$ with Eq. (18);
\STATE Add $\mathcal{X}$ with frequency $s(\mathcal{X})$ into $\mathcal{D}^{rs}$;
\STATE Compute the loss $\mathcal{L}_{total}$ following Eq. (17);
\STATE Update parameters $\theta$ with dataset $\mathcal{D}^{rs}$;
\ENDIF
\ENDFOR
\ENDFOR
\end{algorithmic}
\end{algorithm}

\begin{table}[t]
    \centering
    \begingroup
    \setlength{\tabcolsep}{4.5pt} % Default value: 6pt
    \renewcommand{\arraystretch}{1} % Default value: 1

    \begin{tabular}{ c  c  c  c}
    \toprule
    \textbf{Model} & KS &UCF-51 & Food-101 \\ \midrule \midrule

    Random re-sample &$60.78$ &$68.59$ &$84.21$ \\
    Inverse re-sample &$57.24$ &$66.19$ &$78.26$ \\
    DSR ($k=-0.5$) &$63.68$ &$72.49$ &$89.76$ \\
    DSR ($k=-1.0$) &$64.33$ &$74.11$ &$91.28$ \\
    DSR ($k=-1.5$) &$66.03$ &$73.58$ &$91.74$ \\
    DSR ($k=-2.0$) &\textcolor{blue}{$66.52$} &$75.60$ &$93.36$ \\
    DSR ($k=-2.5$) &$66.35$ &\textcolor{blue}{$76.27$} &\textcolor{blue}{$93.55$} \\
    DSR ($k=-3.0$) &\textcolor{red}{$67.41$} &\textcolor{red}{$76.83$} &\textcolor{red}{$93.69$} \\
    \bottomrule
    \end{tabular}
    \endgroup
    \vspace{-3pt}
    \caption{Comparison with different re-sample frequencies. $k$ represents the slope of the sampling function $\mathcal{F}_s$, where red and blue indicates the \textcolor{red}{best}/\textcolor{blue}{runner-up} performance. }

    \label{tab_resample}

\end{table}

\noindent The whole training pipeline is provided in Algorithm \ref{alg:algorithm}.
ARM outlines a contribution enhancement strategy designed to address imbalanced multimodal learning challenges. 
It begins by initializing $\mathcal{D}^{rs}$ as $\mathcal{D}$. For each sample of multimodal inputs $\mathcal{X} = \{ x^1, x^2, \dots, x^m \}$ in $\mathcal{D}^{rs}$, ARM evaluates the multimodal joint contribution first. After the warm-up period, the core of the algorithm activates, and the contribution scores guide a dynamic feature-level fusion process. Next, the algorithm calculates the re-sampling frequency $s(\mathcal{X})$ for each sample and updates $\mathcal{D}^{rs}$ based on these frequencies. The process then repeats loss $\mathcal{L}_{total}$ computation and parameter updates using the modified dataset.
This approach enhances learning by continuously adjusting contributions from each modality based on their mutual interactions, promoting balanced learning in multimodal scenarios.

\section{C. More Discussions}

\subsection{Analysis of Different Re-sample Frequency}
In this section, we provide a comparison of the results with other re-sample methods and different sampling frequencies to answer \textbf{Q5:} \textit{Does our dynamic sample-level re-sample strategy really work?}

We compare with two related re-sample settings, Random re-sample is to randomly re-sample input of each sample with the same frequency, Inverse re-sample is only resampling the sample with higher contribution. Our proposed dynamic sample-level re-sample (DSR), sampling function $\mathcal{F}_s= round<k\phi^{CMI}(\mathcal{X})-km>$, where $\mathcal{X}$ represents a sample, $k$ is the slope of function and $m$ is the number of modality types, $round<>$ represents rounding operation.

Table \ref{tab_resample} compares various re-sampling strategies, focusing on our proposed DSR method with different slopes $k$ for the sampling function $\mathcal{F}_s$. The slope $k$ determines the sampling frequency, where smaller $k$ values indicate higher re-sampling frequencies.
In terms of performance, DSR with varying $k$ values consistently outperforms random and inverse re-sampling strategies. On the KS dataset, the best performance is achieved with $k=-3.0$, yielding an accuracy of 67.41$\%$, which is 6.63$\%$ higher than the random re-sampling baseline. Similarly, for UCF-51 and Food-101, the optimal $k$ values lead to accuracy improvements of 8.24$\%$ and 9.48$\%$, respectively.
The results highlight the advantages of our DSR method, particularly its ability to dynamically adjust sampling based on modality contribution. DSR effectively balances the sampling frequency, ensuring that critical samples are revisited more frequently while avoiding over-sampling less informative samples. This dynamic approach allows for more efficient learning, leading to substantial performance improvements.

However, as the sampling frequency increases (i.e., as $k$ decreases), the computational cost also grows. This is because more frequent re-sampling leads to higher data processing requirements, which may limit the scalability of the approach in large-scale applications. Additionally, while increased sampling frequency can boost performance, there is a performance ceiling. For example, in UCF-51 and Food-101, the performance gains plateau as $k$ decreases from $-2.0$ to $-3.0$. This indicates that beyond a certain point, further increasing the sampling frequency yields diminishing returns. In summary, DSR provides a robust and flexible re-sampling strategy that outperforms traditional methods by dynamically adjusting to modality importance. However, it is important to balance the trade-off between sampling frequency and computational efficiency, as well as to recognize that the performance gains have practical limits.

\subsection{More Ablation Study}
We conducted ablation studies on the three modules in ARM, i.e. dynamic feature-level fusion (DFF), balanced min-max loss (BMML) and dynamic sample-level re-sample (DSR), to answer \textbf{Q6:} \textit{How much does each module contribute in ARM?}

\begin{table}[t]
    \centering
    \begingroup
    \setlength{\tabcolsep}{5pt} % Default value: 6pt
    \renewcommand{\arraystretch}{1} % Default value: 1

    \begin{tabular}{ c  c  c c c c}
    \toprule
    DFF & BMML & DSR &KS &UCF-51 &Food-101 \\ \midrule \midrule
     & & & $59.61$ &$68.23$ &$82.38$\\
    \checkmark & & & $64.34$ &$72.84$ &$91.29$\\
     &\checkmark & & $63.65$ &$72.68$ &$90.83$\\
     & & \checkmark& $63.78$ &$71.81$ &$90.57$\\
    \checkmark &\checkmark & & $65.09$ &\textcolor{blue}{$74.29$} &\textcolor{blue}{$92.40$}\\
    \checkmark & &\checkmark & \textcolor{blue}{$65.21$} &$73.05$ &$91.87$\\
     &\checkmark &\checkmark & $64.58$ &$73.76$ &$92.16$\\
    \checkmark &\checkmark &\checkmark & \textcolor{red}{\textbf{$66.52$}} &\textcolor{red}{\textbf{$75.60$}} &\textcolor{red}{\textbf{$93.36$}} \\
    \bottomrule
    \end{tabular}
    \endgroup
    \caption{Ablation study of each component in ARM.}
    \label{tab_ab}
\end{table}

Table \ref{tab_ab} presents the ablation study results, which highlight the contributions of each ARM component. The individual effects of DFF, BMML, and DSR demonstrate how each component impacts performance.
When DFF is applied alone, the accuracy on the KS dataset improves from 59.61$\%$ to 64.34$\%$, and similar gains are observed on UCF-51 and Food-101 datasets. DFF enhances feature interactions by dynamically adjusting the contributions of different modalities, allowing better feature fusion and synergy. This improvement illustrates how capturing the complementary information between modalities boosts overall performance.
Incorporating BMML further boosts accuracy. All dataset sees an increase. BMML mitigates the impact of imbalanced contributions by balancing the influence of dominant and weaker modalities, which is crucial in real-world scenarios where some modalities may naturally dominate.
The addition of DSR produces significant performance gains across all datasets. For example, on UCF-51, the accuracy jumps to 74.29$\%$, while Food-101 reaches 92.40$\%$. DSR dynamically adjusts the sampling frequency based on each modality’s marginal contribution, ensuring that underrepresented modalities receive adequate focus during training. This dynamic resampling mechanism enhances model robustness, particularly when data distribution is skewed or modalities vary in importance.

When all three components: DFF, BMML, and DSR are integrated, the model achieves the highest performance across all datasets. The combined benefits stem from comprehensive enhancement strategies: better feature fusion, balanced contribution, and dynamic sample reweighting. The performance boost demonstrates that the integration of these components synergistically addresses the challenges of modality imbalance, feature misalignment, and suboptimal sampling.
In summary, each ARM component provides distinct advantages, with DFF improving feature alignment, BMML addressing modality imbalance, and DSR optimizing sampling. Their combined impact leads to superior accuracy, reflecting their effectiveness in enhancing multimodal learning.

\subsection{Results on Other Fusion Methods} 
Notably, our method is not limited to fixed imbalanced multimodal learning frameworks; it can also be integrated into other existing approaches. In this section, we answer \textbf{Q7:} \textit{how our model improves the performance of other multimodal fusion learning frameworks?}

\begin{table}[t]
    \centering
    \begingroup
    \setlength{\tabcolsep}{2pt} % Default value: 6pt
    \renewcommand{\arraystretch}{1} % Default value: 1

    \begin{tabular}{ l  l  l  l}
    \toprule
    \textbf{Model} & KS &UCF-51 & Food-101 \\ \midrule \midrule

    Concatenation	&$59.61$ &$68.23$	&$82.38$\\
    Concatenation-ARM &$66.52$\scriptsize{\textcolor{green}{$\Delta6.91$}} &$75.60$\scriptsize{\textcolor{green}{$\Delta7.37$}} &$93.36$\scriptsize{\textcolor{green}{$\Delta10.98$}}\\ \midrule
    Summation	&$59.53$ &67.62	&$82.63$\\ 
    Summation-ARM &$66.03$\scriptsize{\textcolor{green}{$\Delta6.50$}} &$74.12$\scriptsize{\textcolor{green}{$\Delta6.50$}} &$93.88$\scriptsize{\textcolor{green}{$\Delta11.25$}}\\ \midrule
    MMTM &$63.92$ &$70.21$ &$90.63$\\
    MMTM-ARM &$67.43$\scriptsize{\textcolor{green}{$\Delta3.51$}} &$75.92$\scriptsize{\textcolor{green}{$\Delta5.71$}} &$94.69$\scriptsize{\textcolor{green}{$\Delta4.06$}}\\ \midrule
    CentralNet &$64.58$ &$72.21$ &$90.31$\\
    CentralNet-ARM &$68.78$\scriptsize{\textcolor{green}{$\Delta4.20$}} &$76.30$\scriptsize{\textcolor{green}{$\Delta4.09$}} &$94.85$\scriptsize{\textcolor{green}{$\Delta4.54$}}\\
    \bottomrule
    \end{tabular}
    \endgroup
    \caption{Results of using ARM on various multimodal fusion Methods, \textcolor{green}{$\Delta$} is accuracy enhancement.}

    \label{tab_enhance}

\end{table}

Table \ref{tab_enhance} illustrates the performance improvement achieved by integrating ARM into various multimodal frameworks, including MMTM, and CentralNet.
For the KS dataset, ARM consistently boosts accuracy across all frameworks, with improvements ranging from 6.91$\%$ for Concatenation to 4.20$\%$ for CentralNet. On other dataset, ARM shows similar trends.
The substantial performance gains can be attributed to ARM's design, which dynamically balances the contributions of each modality. By addressing cross-modal biases, ARM prevents the dominance of any single modality and ensures more holistic learning. Additionally, ARM’s focus on multimodal fusion allows it to capture complex relationships, effectively leveraging information across modalities. The consistent improvements across different architectures validate ARM's robustness and demonstrates its ability to adapt to various multimodal scenarios.

\begin{table}[t]
    \centering
    \begingroup
    \setlength{\tabcolsep}{4pt} % Default value: 6pt
    \renewcommand{\arraystretch}{1} % Default value: 1

    \begin{tabular}{ l  c  c  c}
    \toprule
    \textbf{Model} & KS &UCF-51 & Food-101 \\ \midrule \midrule

    Local MI \scriptsize{(MICCAI 2021)}	&$61.25$ &$69.82$	&$85.77$ \\
    MI-Dependency \scriptsize{(EMNLP 2021)} &$59.83$ &$67.43$ &$85.24$ \\
    Infomax \scriptsize{(EMNLP 2021)} &$62.54$ &$70.04$ &$89.26$ \\
    AMID \scriptsize{(CVPR 2023)} &\textcolor{blue}{$64.73$} &\textcolor{blue}{$73.80$} &\textcolor{blue}{$90.18$} \\
    ARM &\textcolor{red}{$66.52$} &\textcolor{red}{$75.60$} &\textcolor{red}{$93.36$} \\
    \bottomrule
    \end{tabular}
    \endgroup
    \vspace{-3pt}
    \caption{Comparison with Mutual information (MI)-based multimodal learning methods.}

    \label{tab_mi}

\end{table}

\begin{figure}[t]
    \centering
    \includegraphics[width=0.9\linewidth]{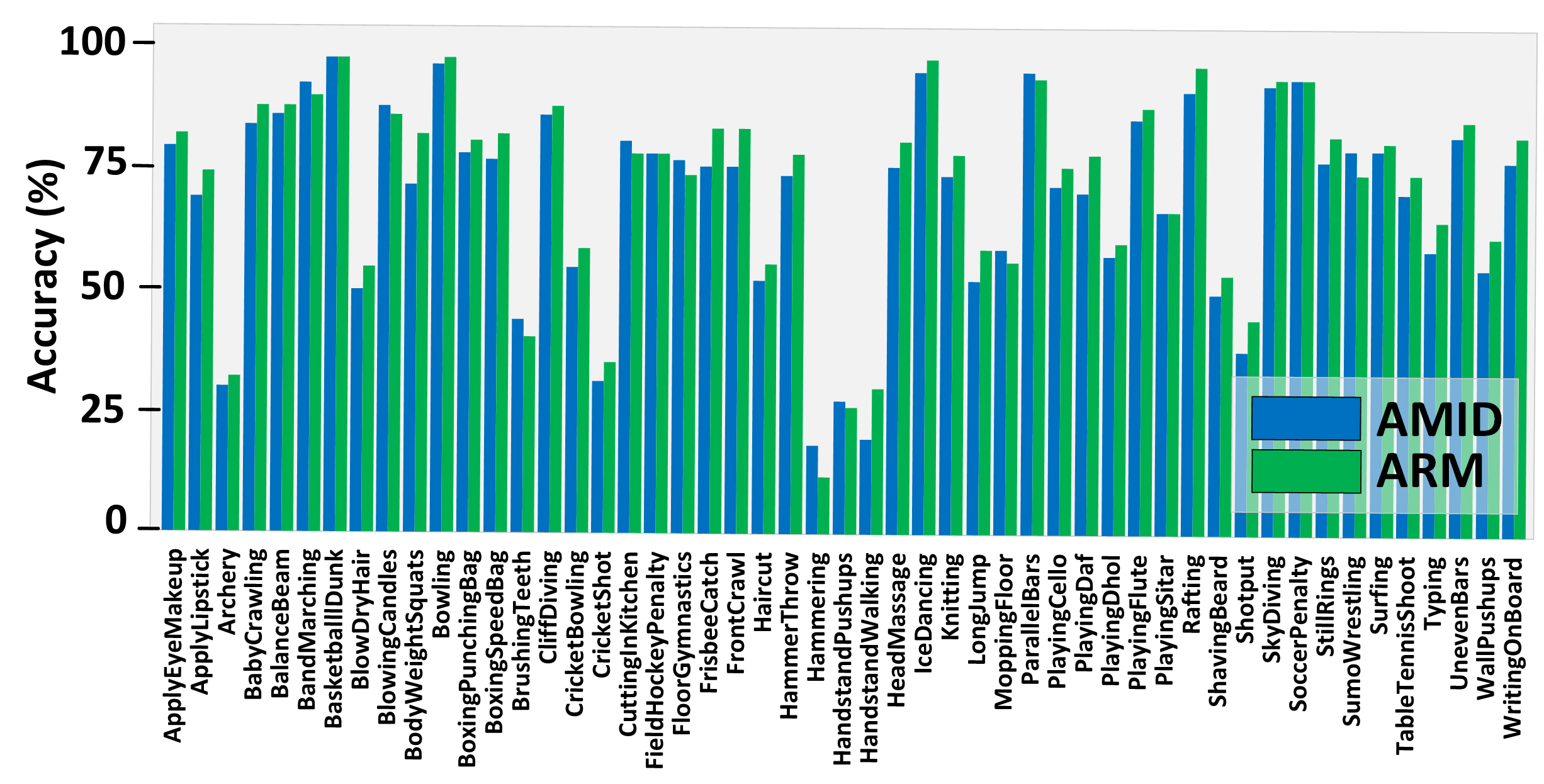}
    \caption{ The per-class accuracy ($\%$) of recognition on UCF-51 comparing \textbf{ARM} with AMID.}
    \label{fig_per}
\end{figure}

\begin{figure*}[t]
    \centering
    \includegraphics[width=0.9\linewidth]{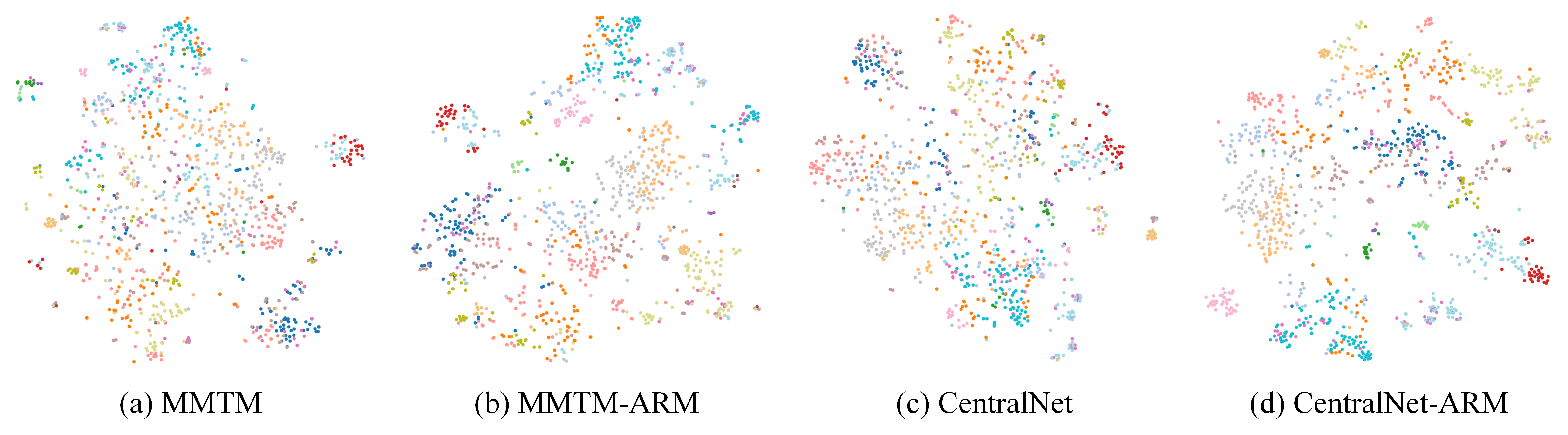}
    \caption{Visual feature distribution of MMTM, MMTM-ARM and CentralNet, CentralNet-ARM visualized by t-SNE \cite{tsne} on Kinetics Sounds dataset. Categories are indicated in different colors.}
    \label{fig-tsne}
\end{figure*}

Visualizations in Fig. \ref{fig-tsne} compare the feature space distributions of MMTM and CentralNet, before and after integrating our proposed ARM method.
Without ARM, both MMTM and CentralNet display noticeable overlaps between clusters, indicating less discriminative feature spaces. After incorporating ARM, the cluster separations become more distinct and well-defined, suggesting improved feature representation and class separability. Specifically, ARM reduces intra-class variance and enhances inter-class separability, resulting in more cohesive clusters with minimal scatter. This enhancement translates into better classification performance.
By addressing the imbalances in multimodal fusion, ARM not only strengthens the learning process for underrepresented modalities but also refines the overall joint feature space, leading to superior cluster organization. These qualitative improvements highlight how ARM effectively amplifies the strengths of existing multimodal networks like MMTM and CentralNet, demonstrating its general applicability and effectiveness in diverse multimodal scenarios.
\begin{figure*}[t]
    \centering
    \includegraphics[width=0.8\linewidth]{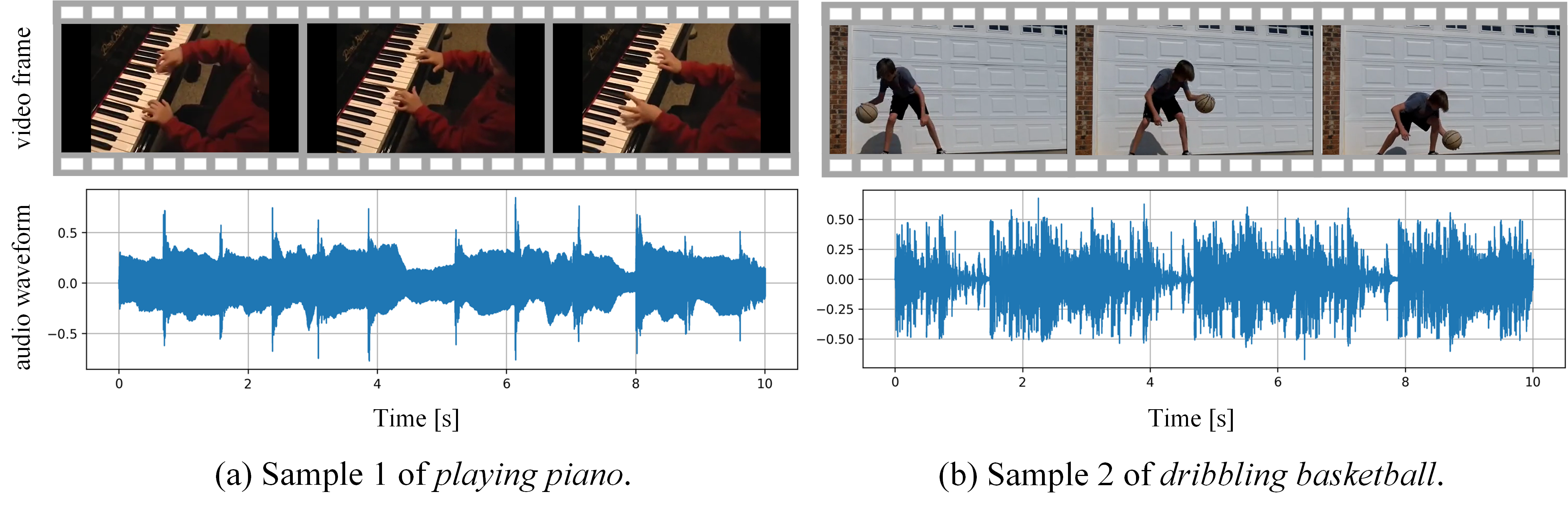}
    \caption{Visualization of Audio-visual samples from Kinetics Sounds dataset.}
    \label{fig-sample}
\end{figure*}

\subsection{More Comparisons}
In this section, we answer \textbf{Q8:} \textit{How much does our method improve performance compared to existing MI-based multimodal learning methods?}

Table \ref{tab_mi} presents a performance compared with other mutual information (MI)-based multimodal learning methods, including Local-MI \cite{localmi}, Mutual-Dependency \cite{midepend}, Multimodal-Infomax \cite{infomax} and AMID \cite{kd-mi}. Our ARM outperforms the competing approaches by a significant margin on all three datasets. 
The notable performance gap highlights the effectiveness of ARM in capturing and balancing the complex multimodal relationships, which are critical for robust feature integration. 
The superior performance of ARM can be attributed to dynamically adjust contributions from each modality, allowing ARM address modality imbalance and information redundancy better, which are common issues in MI-based methods.

The per-class accuracy comparison between ARM and AMID in Fig. \ref{fig_per} demonstrates ARM's consistent performance advantage across various fine-grained action categories. While both methods show competitive results, ARM achieves superior accuracy in a majority of categories. This indicates that ARM is more effective in capturing the subtle nuances and intricate features within multimodal inputs, leading to better classification outcomes.
Additionally, in categories where AMID struggles with lower accuracy, such as \textit{Hand stand Pushups} and \textit{HammerThrow}, ARM maintains a stable and high performance, reflecting its robustness in handling challenging and diverse actions. This consistency across the spectrum suggests that ARM effectively mitigates modality imbalance and enhances joint learning across different classes. Overall, the detailed analysis of fine-grained categories highlights ARM's strength in generalizing across varied scenarios while delivering more balanced and reliable results than AMID.

\subsection{Case Analysis of Modality Contribution}
Here we provide a visualization instance to answer \textbf{Q9:} \textit{How does ARM balance two modalities in samples with different contributions?}

Fig. \ref{fig-sample} show two audio-visual multimodal pair of \textit{playing piano} and \textit{dribbling basketball} category, respectively. In Sample 1, the clear piano sound in the audio modality is easily recognizable, while the bouncing basketball action in Sample 2 is hard to detect due to unrelated background music interference. This could drag the joint contribution of all modalities by the more challenging-to-learn modality.

\begin{figure*}[ht]
    \centering
    \includegraphics[width=0.9\linewidth]{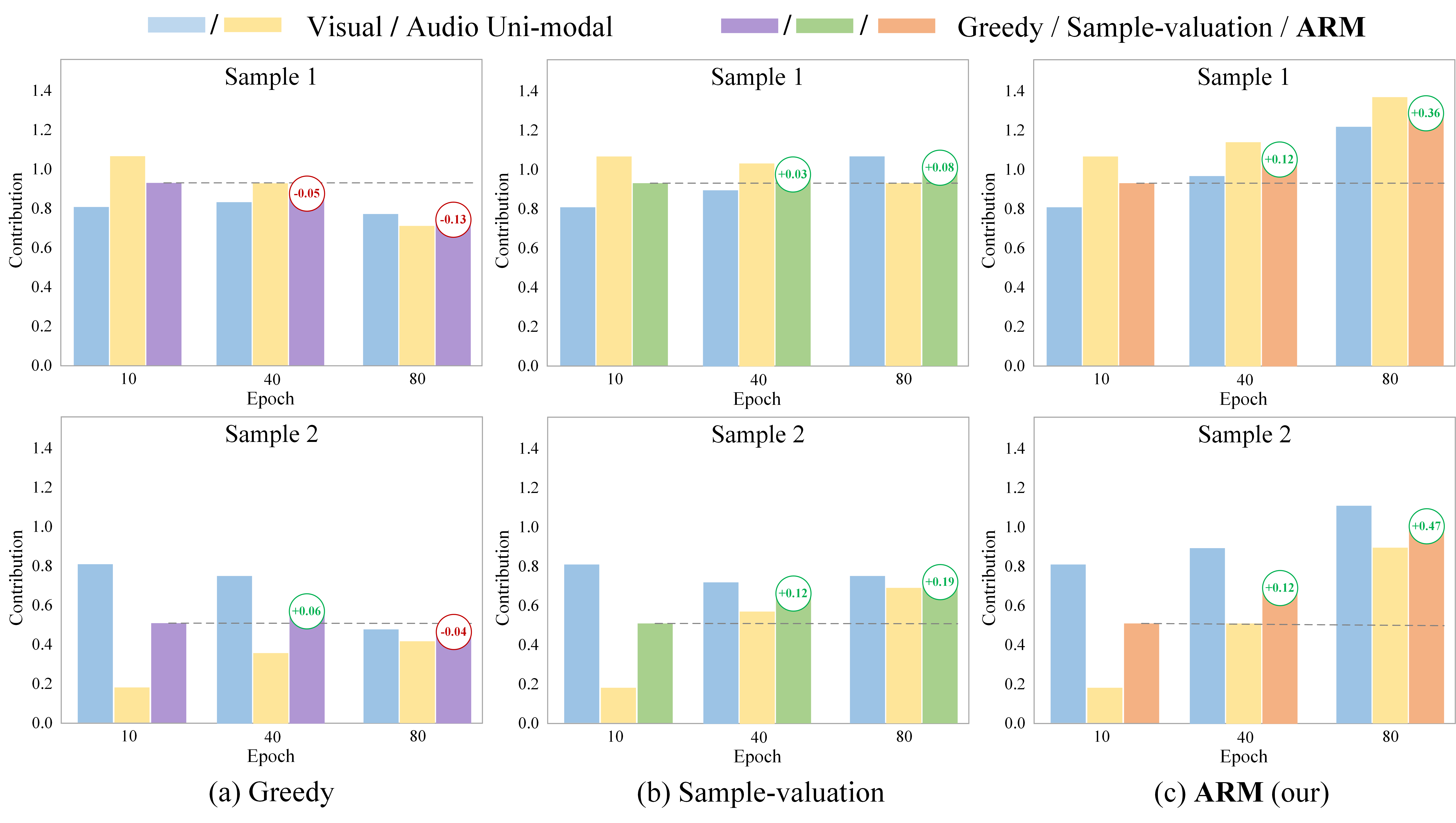}
    \caption{Contribution improvement compared. Other imbalanced multimodal learning methods: Greedy \cite{greedy}, Sample-valuation \cite{sample}.}
    \label{fig-contr}
\end{figure*}

Fig. \ref{fig-contr} compare contribution improvement for this two audio-visual samples under different imbalanced multimodal learning methods: Greedy, Sample-valuation, and our ARM.
The contribution of each modality is tracked across training epochs (10, 40, 80). The results show that both Greedy and Sample-valuation exhibit fluctuating and imbalanced contributions, the focus on lower-contribution modality often leads to a decrease in higher-contribution modality. Meanwhile, they fail to maintain consistent contributions across epochs, resulting in instability.
In contrast, our ARM demonstrates balanced and stable contribution enhancement from all modalities. The key reason for this advantage lies in ARM’s dynamic re-sampling strategy and balanced min-max loss, which adaptively adjust the sampling frequency and contribution of each modality based on their marginal and joint contributions. This ensures that neither modality is overemphasized or ignored, leading to better generalization and more robust feature fusion. Consequently, ARM is able to achieve superior performance in scenarios with imbalanced modalities by maintaining consistent contribution levels, helping in maximizing the joint contribution gain.

\end{document}